\documentclass[journal,onecolumn, draftclsnofoot]{IEEEtran}
\usepackage{graphicx}%
\usepackage{epsfig}
\usepackage{graphicx}%
\usepackage{multirow}%
\usepackage{amsmath,amssymb,amsfonts}%
\usepackage{amsthm}%
\usepackage{mathrsfs}%
\usepackage{xcolor}%
\usepackage{textcomp}%
\usepackage{manyfoot}%
\usepackage{booktabs}%
\usepackage{algorithm}%
\usepackage{algorithmicx}%
\usepackage{algpseudocode}%
\usepackage{listings}%
\usepackage[flushleft]{threeparttable}

\usepackage{setspace}
\begin{document}
\title{ExtremeMETA: High-speed Lightweight Image Segmentation Model by Remodeling Multi-channel Metamaterial Imagers}

\author{Quan Liu, Brandon T. Swartz, Ivan Kravchenko, Jason G. Valentine, Yuankai Huo,% <-this % stops a space
\thanks{Yuankai Huo is the corresponding author, Vanderbilt University, Nashville, TN 37212, USA. E-mail: (yuankai.huo@vanderbilt.edu)}
\thanks{Quan Liu, Brandon T. Swartz and Jason G. Valentine are with the Vanderbilt University, Nashville, TN 37212, USA. E-mail: (quan.liu@vanderbilt.edu; brandon.t.swartz@vanderbilt.edu; jason.g.valentine@vanderbilt.edu).}% <-this % stops a space
\thanks{Ivan Kravchenko is with Oak Ridge National Laboratory, Oak Ridge, TN 37830, USA. E-mail: (kravchenkoii@ornl.gov)}% <-this % stops a space
}

\maketitle

\begin{abstract}
Deep neural networks (DNNs) have heavily relied on traditional computational units like CPUs and GPUs. However, this conventional approach brings significant computational burdens, latency issues, and high power consumption, limiting their effectiveness. This has sparked the need for lightweight networks like ExtremeC3Net. On the other hand, there have been notable advancements in optical computational units, particularly with metamaterials, offering the exciting prospect of energy-efficient neural networks operating at the speed of light. Yet, the digital design of metamaterial neural networks (MNNs) faces challenges such as precision, noise, and bandwidth, limiting their application to intuitive tasks and low-resolution images.
In this paper, we propose a large kernel lightweight segmentation model, ExtremeMETA. Based on the ExtremeC3Net, the ExtremeMETA maximizes the ability of the first convolution layer by exploring a larger convolution kernel and multiple processing paths. With the proposed large kernel convolution model, we extend the optic neural network application boundary to the segmentation task. To further lighten the computation burden of the digital processing part, a set of model compression methods is applied to improve model efficiency in the inference stage. The experimental results on three publicly available datasets demonstrate that the optimized efficient design improved segmentation performance from 92.45 to 95.97 on mIoU while reducing computational FLOPs from 461.07 MMacs to 166.03 MMacs. The proposed the large kernel lightweight model ExtremeMETA showcases the hybrid design's ability on complex tasks.

\end{abstract}


% \fi
% \IEEEpeerreviewmaketitle

\section{Introduction}

In the realm of modern computer vision, digital neural networks play a pivotal role. The convolutional neural network (CNN) stands out as arguably the most extensively employed AI approach, particularly in tasks like image classification, segmentation, and detection. Despite the advent of vision transformer-based models, convolution remains integral for extracting local image features. Presently, CNNs are typically implemented on computational units like CPUs and GPUs. However, this conventional design approach brings forth substantial challenges, including a formidable computational load, notable latency issues, and heightened power consumption. These limitations become particularly pronounced in applications such as the Internet of Things (IoT), edge computing, and drone operations, where requires the lightweight model for efficient analysis. Recognizing the critical need for DNN models with reduced energy consumption and lower latency, the AI community has embarked on a quest for more efficient solutions. Despite these efforts, achieving energy-free and light-speed DNNs within the current research trends seems to be an elusive goal. 

\begin{figure*}
\begin{center}
\includegraphics[width=0.95 \linewidth]{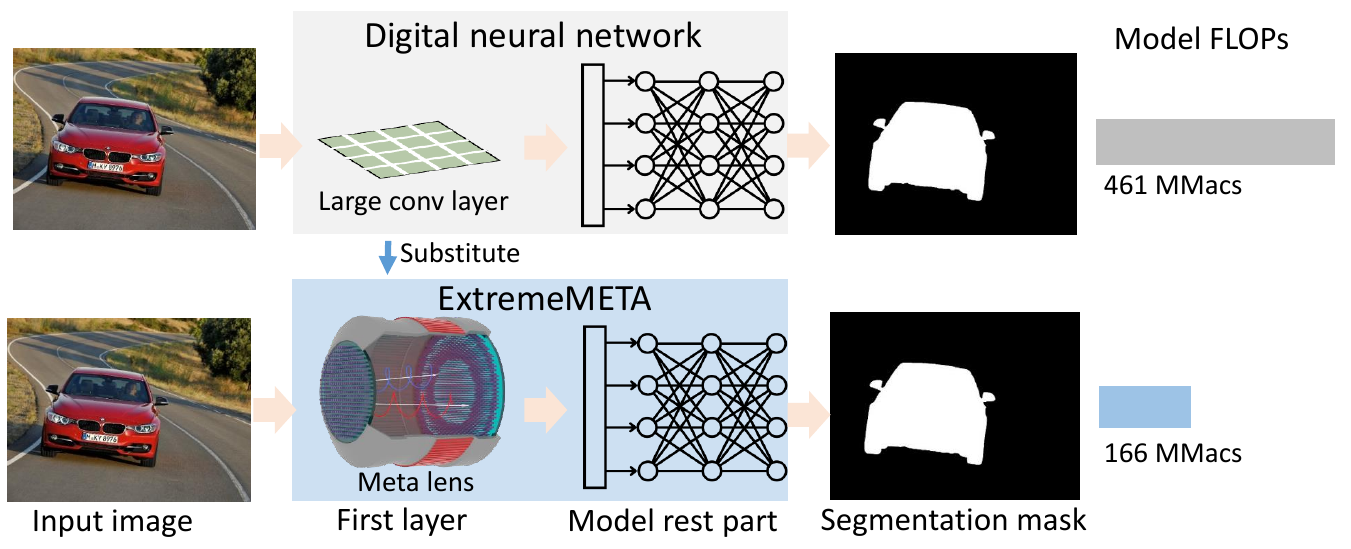}
\end{center}
   \caption{This study provides a hybrid pipeline for designing and optimizing a large kernel digital neural network. The proposed
    ExtremeMETA is efficient for segmentation tasks with less FLOPs in computation.}
\label{fig:idea}
 \end{figure*}

Recent breakthroughs in optical computational units, including metamaterials (refer to Fig.~\ref{fig:idea}), have brought to light the potential for neural networks that operate without energy consumption and at unprecedented speeds. The current cutting-edge metamaterial neural network (MNN) takes on a hybrid form, leveraging optical processors as a light-speed and energy-free front-end convolutional operator alongside a digital feature aggregator. This inventive approach serves to significantly reduce computational latency. By assigning the convolution operations to optical units, more than 90 percent of the floating-point operations (FLOPs) inherent in conventional CNN backbones like VGG and ResNet are effectively off-loaded. This marks a noteworthy departure from traditional architectures, opening up new avenues for efficient and high-performance neural network designs. However, the hybrid design is fundamentally influenced by the physical structure including the limited kernel size, and channel number. Besides that, the hybrid system is also limited by what can be fabricated as the first optical layer of the neural network. 

Based on our proposed LMNN model, the hybrid design achieves promising performance on the classification task. However, there are limitations of LMNN, namely: (1) the LMNN model can only perform image classification tasks instead of model complex tasks like image segmentation and object detection; (2) input images are in low resolution (28$\times$28), and (3) besides leverage the computation burden to the optic, the digital part requires efficiency improvement operation like model compression in the inference stage. 

In this paper, we propose a novel large kernel lightweight segmentation model ExtremeMETA that maximizes the efficiency advantages of optic signal computation, while also compressing the digital processing model to further improve the model segmentation efficiency. To adapt the segmentation task on large images, the proposed lightweight large kernel model achieves larger receptive fields, the ability to larger image analysis, and covers general vision tasks, image classification segmentation, and detection. Furthermore, the complexity of the model digital processing part is explicitly addressed via a set of model compression methods. We evaluate our design on image segmentation tasks using three public datasets: the portrait dataset, the Stanford dataset, and KITTI dataset. The proposed lightweight large kernel model achieved superior segmentation accuracy as compared with the SOTA segmentation model. Overall, the system's contributions can be summarized in four-fold:

\begin{itemize}
\item We propose a new large convolution kernel CNN network to achieve a large reception field, less energy consumption, and less latency. 
\item We introduce the model re-parameterization to improve large convolution kernel performance and sparse convolution kernel compression mechanism to compress the multi-branch sparse-convolution design to a single layer for the hybrid system implementation. The model compression mechanism improves the model efficiency for digital processing.
\item The task limitations of large convolution hybrid models are explicitly addressed via performing segmentation tasks on multiple datasets from different categories.
\end{itemize}

The rest of the paper is organized as follows. In Section II, we introduce background and related research relevant to large kernel convolution, model compression, and optical neural networks on image processing tasks. In Section III, our proposed lightweight lightspeed model is presented. It includes the large kernel re-parameterization, sparse convolution compression, and multi-path model compression. Section IV focuses on presenting the dataset and experiment implementation details. Section V provides the experimental results and ablation study. Then, in Sections VI and VII, we provide the discussion and conclude our work.

\section{Related work}

\subsection{Large kernel convolution design}
In the realm of convolutional neural networks (CNNs), the design and utilization of large kernel convolutions have garnered significant attention in recent years. Numerous studies have explored the benefits of using larger convolutional kernels, such as 7x7 or 11x11, to capture broader spatial contexts and more intricate patterns within images \cite{simonyan2014very,szegedy2015going}. Early research efforts focused on understanding the impact of kernel size on model performance, with findings suggesting that larger kernels can lead to improved feature extraction and recognition accuracy, especially for complex visual tasks \cite{zeiler2014visualizing}.

Building upon these findings, subsequent works have proposed various strategies to incorporate the large kernel convolutions into CNN architectures effectively. These strategies often involve modifying network architectures, adjusting kernel sizes, or integrating multi-scale features to enhance the robustness and versatility of CNN models \cite{szegedy2016rethinking,he2016deep}. Additionally, advancements in hardware acceleration and parallel processing have facilitated the efficient implementation of large kernel convolutions, enabling their widespread adoption across diverse computer vision applications \cite{zhang2018efficient,sun2018efficient}.

Overall, the related work on large kernel convolution design underscores its pivotal role in advancing the capabilities of CNNs for tackling increasingly complex and demanding visual recognition tasks \cite{lin2013network,huang2017densely}.

\subsection{Optic neural network}

Optic neural networks (ONNs) have emerged as a promising paradigm for accelerating neural network computations by leveraging the unique properties of optical computing. Inspired by the principles of light-based signal processing, ONNs exploit the parallelism, high bandwidth, and low energy consumption inherent in optical systems to achieve significant computational efficiency gains compared to traditional electronic implementations. A considerable body of research has focused on exploring various aspects of ONNs, including optical device design, system architectures, and algorithmic frameworks tailored to optical computing platforms \cite{shen2017deep,lin2018all,hughes2018training}.

Early studies laid the groundwork for ONNs by demonstrating their potential for accelerating matrix-vector multiplications, a fundamental operation in neural network inference \cite{tait2017physics,tait2016optical}. Subsequent works have extended ONN capabilities to encompass more complex neural network layers and architectures, paving the way for practical applications in tasks such as image classification, object detection, and natural language processing \cite{miscuglio2018all,larger2012photonic}.

Key challenges in ONN research include addressing optical noise, device nonlinearity, and scalability issues, which require interdisciplinary efforts spanning optics, photonics, and machine learning \cite{jutamulia1996overview,boehm2022harnessing}. Despite these challenges, ONNs hold great promise for enabling ultra-fast and energy-efficient neural network computations, with the potential to revolutionize various domains of artificial intelligence and computing \cite{zhuge2021kaleido,ovchinnikov1999diffraction}.

\subsection{Convolution neural network model compression}

In the field of convolutional neural networks (CNNs), model compression techniques have garnered significant attention as a means to reduce the computational complexity and memory footprint of deep learning models without sacrificing performance. A diverse range of methods has been proposed to compress CNNs, including pruning, quantization, low-rank approximation, knowledge distillation, and weight sharing. Pruning techniques aim to remove redundant or less important parameters from the network, thereby reducing its size and computational cost \cite{han2015learning,molchanov2016pruning}. Quantization methods reduce the precision of network parameters, often by representing weights and activations with fewer bits, to decrease memory requirements and improve inference speed \cite{hubara2017quantized}. Low-rank approximation techniques exploit the underlying structure of weight matrices to factorize them into smaller, more computationally efficient components \cite{denton2014exploiting}. Knowledge distillation involves training a compact "student" network to mimic the predictions of a larger "teacher" network, transferring knowledge from the latter to the former \cite{hinton2015distilling}. Additionally, weight sharing approaches aim to reduce redundancy by sharing parameters across different parts of the network \cite{chen2015compressing}.

Collectively, these model compression techniques offer effective strategies for deploying CNNs on resource-constrained devices or accelerating inference in large-scale deployment scenarios. Ongoing research in this area continues to explore novel compression algorithms, optimization strategies, and application-specific considerations to further improve the efficiency and effectiveness of compressed CNN models.

\section{Method}
\textbf{Problem statement}
We extensively study the trainability of large kernels on metamaterial neural networks (MNN) and unveil three main observations: (i) traditional convolution kernel shows limited improvement on large images; (ii) the MNN is only available on classification task; (iii) metamaterial implementation limited the computation ratio on segmentation model which is normally in complex structure. 

\begin{figure*}
\begin{center}
\includegraphics[width=0.8 \linewidth]{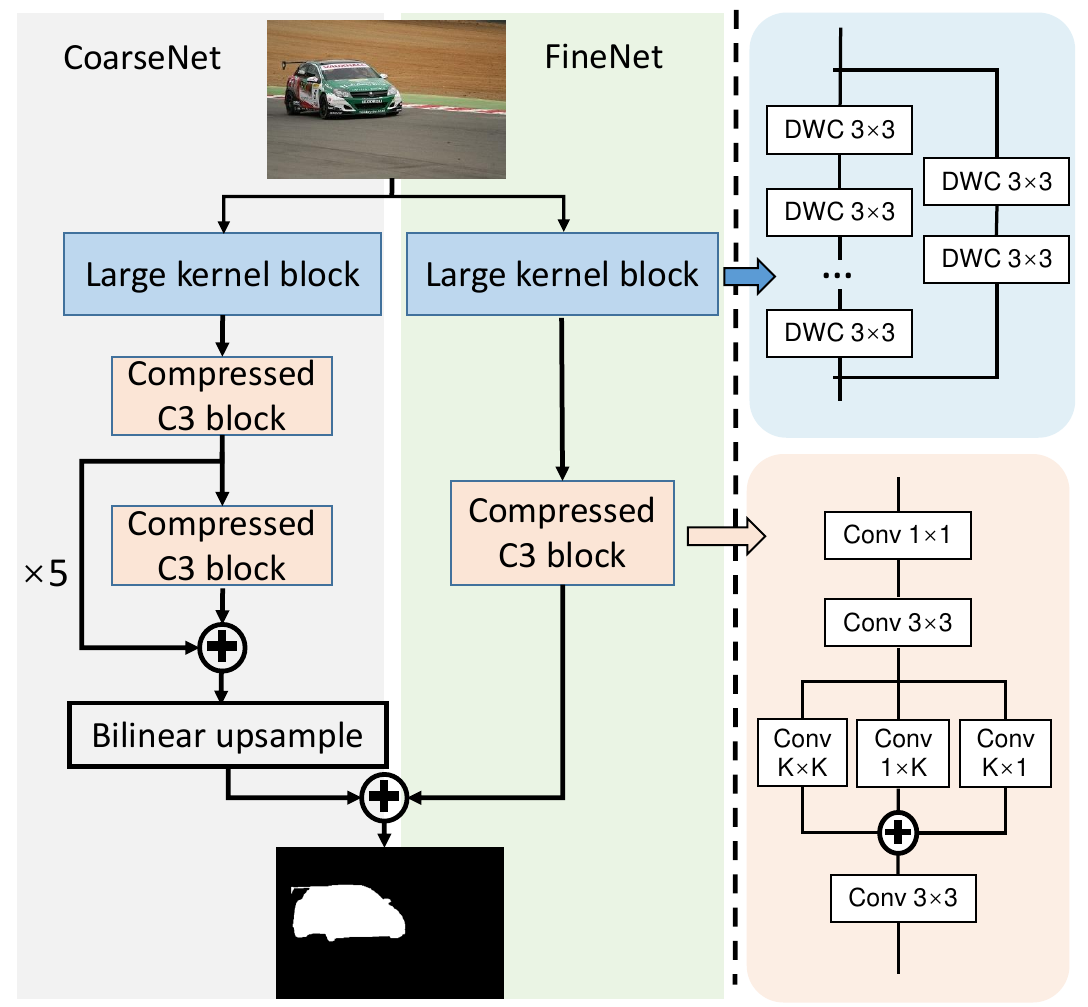}
\end{center}
   \caption{Lightweight segmentation model with hybrid meta optics design. The model has two parts: CoarseNet and FineNet. The large kernel block is composed of depthwise convolution layers.}
\label{fig:pipeline}
 \end{figure*}

\subsection{Large convolution design with multiple path design}
Limited by the image size and the task for the model, our previous proposed model LMNN achieved the prediction performance with kernel size $9 \times 9$. Two major limitations exist when applying the large kernel design to the MNN: (1) the metamaterial implementation limits the image size to a small range; (2) only the classification task is available to be validated on the MNN model when the segmentation task and detection task is too difficult to be implemented under the optic implementation limitation. To address the challenges, we proposed our model from two perspectives: (1) from kernel design, we employ the large convolution kernel with parameterization design to construct the convolution layer (larger than $9\times9$); (2) from model design, our proposed lightweight segmentation model based on the multipath model structure composed by a course segmentation path and a light refinement path proposed by \cite{park2019extremec3net}.

\begin{figure*}
\begin{center}
\includegraphics[width=0.7 \linewidth]{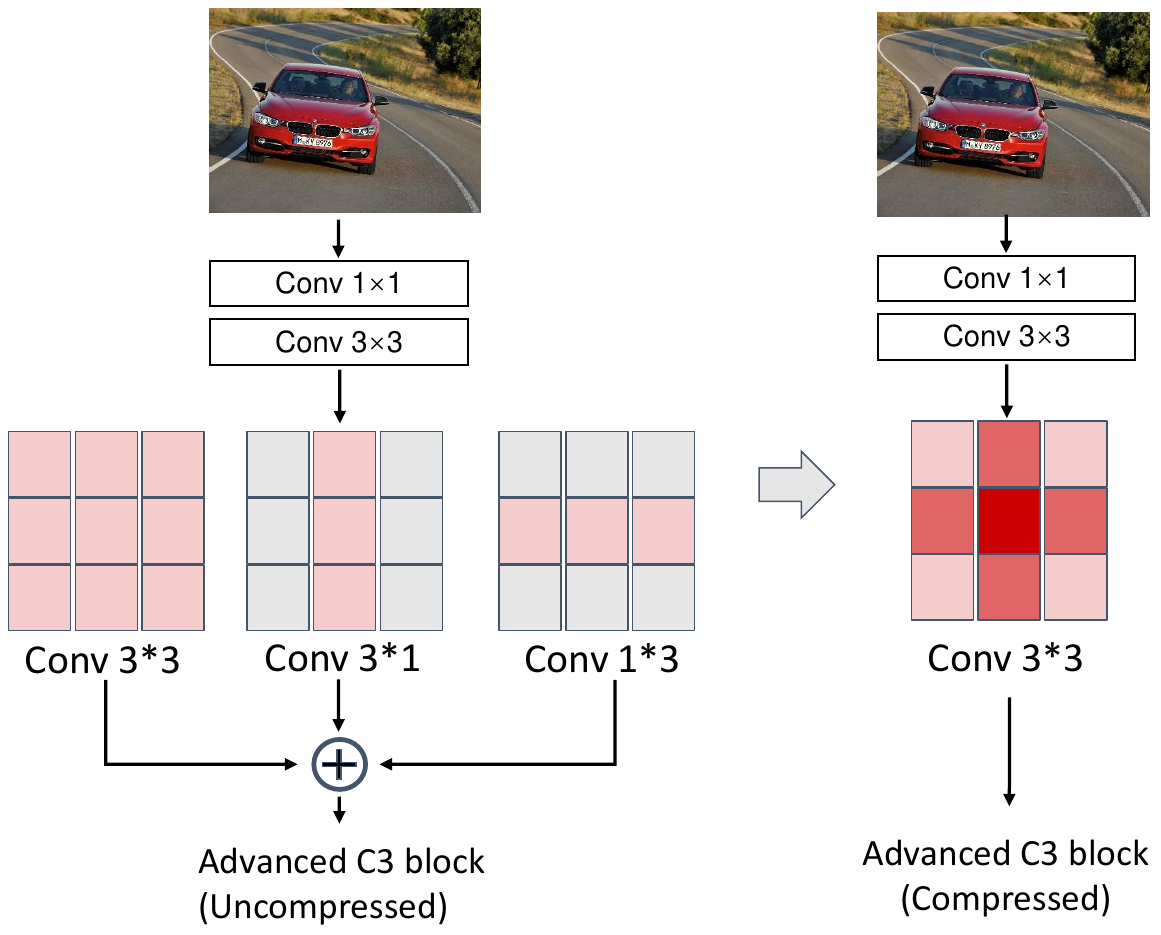}
\end{center}
   \caption{Model compression on segmentation model digital processing part. The left panel shows the multipath structure of the advanced C3 block. The right panel shows the compression mechanism.}
\label{fig:pipeline}
 \end{figure*}

\subsection{Model compression with sparse convolution}
Model compression is a crucial technique aimed at enhancing the efficiency of deep learning models by reducing their size and computational demands while maintaining their performance standards. Among the various strategies employed in model compression, pruning, and quantization stand out as widely adopted methodologies.
Pruning, a prominent technique in model compression, involves the systematic removal of redundant or unnecessary parameters from neural networks. By identifying and eliminating connections that contribute minimally to the model's performance, pruning effectively reduces the model's size and computational requirements. This process allows for a more streamlined network architecture without sacrificing accuracy, making it particularly valuable for resource-constrained environments or deployment on edge devices.

We applied model compression and parameterization together for the sparse convolution kernel which is shown in Fig. Sparse convolution refers to a convolution operation where the kernel (filter) contains mostly zero values, resulting in a sparse structure. When using a kernel size of 1x3 (1 row and 3 columns), the convolution operation typically involves sliding this kernel over the input data and performing element-wise multiplication followed by summation along the spatial dimensions.

\begin{equation}
    O_{h,w,c'} = \sum_{i=0}^{2} \sum_{j=0}^{C-1} I_{h,w+i,j} \times K_{0,i,j,c'}
\end{equation}

I as the input tensor, K as the kernel tensor, O as the output tensor, and $\times$ as the convolution operation.

For the ExtremeC3 block, we have three convolution paths with kernel size $k \times k$, $1 \times k$, and $k \times 1$. The compressed convolution kernel follows Eq. Let's denote the individual kernels as $k_{1 \times k}$, $k_{k \times k}$, $k_{k \times 1}$. 

\begin{equation}
    K_{combined}(i,j) = w_{1xk} \times K_{1xk}(i,j) + w_{kxk} \times K_{kxk}(i,j) + w_{kx1} \times K_{kx1}(i,j)
\end{equation}

The compressed multipath convolution block saves computation complexity in the inference stage.

\section{Data and experimental design}

\subsection{Data description}
Three public datasets, EG1800~\cite{shen2016automatic}, Stanford Car dataset~\cite{krause2013collecting}, and KITTI dataset~\cite{Geiger2012CVPR}, were used to evaluate the lightweight large kernel model on segmentation tasks. For the EG1800 dataset, we employ 1887 images in $600 \times 800$ resolution with semantic segmentation masks. The EG1800 dataset is collected from Flicker with the manually annotated mask of the portrait. The Stanford Car dataset is composed of 16,185 RGB images of cars with the point coordinate where the car is located in images. The KITTI dataset is popular in mobile robotics and autonomous driving and features diverse traffic scenarios captured using high-resolution RGB, grayscale stereo cameras, and a 3D laser scanner. However, it lacks inherent ground truth annotations for semantic segmentation. To adapt to the segmentation task, both the Stanford Car dataset and the KITTI dataset need to address the annotation limitation.

\subsection{Data generation with foundation model}
Regarding the Stanford Car dataset and KITTI dataset lacking of segmentation annotation, we employ the Segment Anything Model (SAM)~\cite{kirillov2023segment} to generate the object mask based on the prompts of object location. The SAM model is a foundation model that has a zero-shot ability to segment objects on new image distributions. The RGB image of Standford Car and KITTI datasets and bounding box coordinate is provided for the SAM model and SAM model will generate the object masks. With the help of the SAM model, the RGB images with object mask annotations are available for model training.  

\subsection{Large kernel digital design on segmentation model}
The large kernel design is applied to the segmentation network's first convolution layer design. Since the first layer is designed to be substituted by the metaoptic lens in the inference stage, our large kernel design is under physical limitation. On the other hand, the optic lens provides light-speed computation which we can take advantage of. 
Based on the multipath segmentation network, the first convolution layers of both the Coarse-net part and Fine-net part are redesigned with the large convolution kernel with parameterization following the strategy in our previous work LMNN~\cite{liu2023digital}. Since the image is large compared with FashionMNIST previously used, our kernel size is larger from $9 \times 9$ to $15 \times 15$. The channel number is expanded from 12 to 48. The Larger convolution kernel and channel number provide the large capability of the first layers and handle the complex situation.

\subsection{Model design with optic constrain}
Under the meta-optic fabrication limitation, the meta-optic layer has limitations on both channel number and input size. The trade-off in model performance between input size and channel number is discussed. The size-first design uses the largest input image size under fabrication constrain. Channel-first design prefers more channel numbers under the fabrication limitation.

\subsection{Model compression efficiency}
Besides enlarging the capability of the first layer, our proposed lightweight segmentation network is compressed in the digital part. Since the model compression affects the model's complexity and efficiency, we evaluate if the compressed model loses accuracy. 
To test the efficiency of the model compression strategy, the model FLOPs, parameters, and FLOPs ratio of the first convolution layer.

\section{Result}
In this section, we evaluate our proposed lightweight segmentation network with a simple model structure, using the EG1800 dataset, Stanford Car dataset, and KITTI dataset. Since the Stanford Car dataset and KITTI dataset are car images, we train the model and test the two datasets together.

\subsection{Segmentation performance on portrait dataset}
We evaluate the lightweight segmentation model on EG1800 dataset together with model parameters and first convolution FLOPs ratio. As shown in Table.~\ref{tab:eg1800_before}, the original ExtremeC3 model cannot take advantage of the large convolution kernel on the first layer, $15 \times 15$ kernel shows even lower performance than $11 \times 11$. The model performance without the first convolution layer shows a 2$\%$ drop compared with the ExtremeC3 model with $3 \times 3$ kernel size. Our proposed hybrid lightweight segmentation model achieves the best performance with $15 \times 15$ convolution kernel which has the same digital computation FLOPs. 

\begin{table}
\centering
% \begin{center}  
\caption{Segmentation performance on EG1800}
% \end{center}
% \centering
\label{tab:eg1800_before}
% \begin{center}   
\begin{tabular}{lccccc}
    \hline
    % \multicolumn{2}{c}{Part}          \\
    % \cmidrule(r){1-2}
        % & \multicolumn{2}{c}{FashionMNIST}   &  \multicolumn{2}{c}{STL-10}  \\
     Model   &  Kernel size  &  1st conv FLOPs ($\%$)  & Model FLOPs & Digital FLOPs & Test (mIoU)  \\
    \hline
    \multirow{3}{*}{ExtremeC3 }   &   3$\times$3    &  10.87  &  199.4  &  199.4  & 0.9249     \\
                 &  11$\times$11   &  62.11  &  469.14 &  469.14  & 0.9323     \\
                 &  15$\times$15   &  75.30  &  719.62 &  719.62  & 0.9301     \\
    \hline
    Digital      &      N/A        &  N/A    &  174.10 &  174.10   & 0.9086     \\
    \hline
    \multirow{4}{*}{Ours}         &   1$\times$1    &  2.80   &  182.06  &  174.10   & 0.9137   \\
                 &   3$\times$3    &  10.87  &  199.40  &  174.10   & 0.9234   \\
                 &  11$\times$11   &  59.68  &  431.81 &  174.10  & 0.9415     \\
                 &  15$\times$15   &  63.36  &  475.16 &  174.10   & \textbf{0.9418}     \\                               
    \hline
\end{tabular}
% \end{center}
\text{Model FLOPs and digital FLOPs unit is MMacs.}
\end{table}

% \begin{table}
% \centering
% % \begin{center}  
% \caption{Segmentation performance on EG1800}
% % \end{center}
% % \centering
% \label{tab:eg1800_before}
% % \begin{center}   
% \begin{tabular}{lcccccc}
%     \hline
%     % \multicolumn{2}{c}{Part}          \\
%     % \cmidrule(r){1-2}
%         % & \multicolumn{2}{c}{FashionMNIST}   &  \multicolumn{2}{c}{STL-10}  \\
%      Model   &  Kernel size  &  1st conv FLOPs ($\%$)  & Model FLOPs & Digital FLOPs & Params (k) & Test (mIoU)  \\
%     \hline
%     \multirow{3}{*}{ExtremeC3 }   &   3$\times$3    &  10.87  &  199.4  &  199.4  & 36.54  & 0.9249     \\
%                  &  11$\times$11   &  62.11  &  469.14 &  469.14 & 41.91  & 0.9323     \\
%                  &  15$\times$15   &  75.30  &  719.62 &  719.62 & 46.91  & 0.9301     \\
%     \hline
%     Digital      &      N/A        &  N/A    &  174.10 &  174.10 & 36.24  & 0.9086     \\
%     \hline
%     \multirow{4}{*}{Ours}         &   1$\times$1    &  2.80   &  182.06  &  174.10  & 36.33  & 0.9137   \\
%                  &   3$\times$3    &  10.87  &  199.40  &  174.10  & 36.54  & 0.9234   \\
%                  &  11$\times$11   &  59.68  &  431.81 &  174.10 & 42.03  & 0.9415     \\
%                  &  15$\times$15   &  63.36  &  475.16 &  174.10 & 42.03  & \textbf{0.9418}     \\                               
%     \hline
% \end{tabular}
% % \end{center}
% \text{Model FLOPs and digital FLOPs unit is MMAC.}
% \end{table}

Besides improving the model performance with advanced design on the first convolution layer, we evaluate the model efficiency improvement by model compression. Following the experiment setting in Table.~\ref{tab:eg1800_before}, we applied model compression, including sparse convolution kernel compression and multipath parameterization, to each model design and shows the efficiency evaluation matrix in Table.~\ref{tab:eg1800_after}. The compression method shows efficient computation on digital FLOPs without affecting model performance (mIoU).

\begin{table}
\centering
% \begin{center}  
\caption{Segmentation performance on EG1800 after model compression}
% \end{center}
% \centering
\label{tab:eg1800_after}
% \begin{center}   
\begin{tabular}{lcccccc}
    \hline
    % \multicolumn{2}{c}{Part}          \\
    % \cmidrule(r){1-2}
        % & \multicolumn{2}{c}{FashionMNIST}   &  \multicolumn{2}{c}{STL-10}  \\
     Model   &  Kernel size  &  1st conv FLOPs ($\%$)  & Model FLOPs & Digital FLOPs & Test (mIoU)  \\
    \hline
    \multirow{3}{*}{ExtremeC3}    &   3$\times$3    &  11.33  &  191.32  &  191.32  & 0.9233   \\
                 &  11$\times$11   &  63.21  &  461.07 &  461.07  & 0.9315     \\
                 &  15$\times$15   &  76.16  &  711.55 &  711.55   & 0.9289     \\
    \hline
    Digital      &      N/A        &  N/A    &  166.03 &  166.03   & 0.9031     \\
    \hline
    \multirow{4}{*}{Ours}         &   1$\times$1    &  3.17   &  174.25  &  166.03  & 0.9121   \\
                 &   3$\times$3    &  11.33  &  191.32  &  166.03  & 0.9217    \\
                 &  11$\times$11   &  60.81  &  423.74 &  166.03  & 0.9404     \\
                 &  15$\times$15   &  64.45  &  467.09 &  166.03 & \textbf{0.9420}     \\                               
    \hline
\end{tabular}
% \end{center}
\text{Model FLOPs and digital FLOPs unit is MMacs.}
\end{table}

\subsection{Segmentation performance on car dataset}
To validate our lightweight segmentation model with more datasets, we conduct experiments on the car dataset, including the Stanford Car dataset and KITTI dataset both with semantic segmentation mask as ground truth. Since the Stanford Car dataset and KITTI dataset are in different resolutions. Both the Standford Car dataset and the KITTI dataset are used for model training.

\begin{table}
\centering  
\caption{Segmentation performance on car dataset}
\label{tab:car_before}
\begin{tabular}{lccccc} \hline Model & Kernel size & Train (KITTI+Stanford) & Test (mIoU) & KITTI & Stanford \\ 
\hline
\multirow{3}{*}{ExtremeC3}   & 3*3 & 95.02 & 92.51 & 84.45 & 95.23 \\ 
            & 11*11 & 95.12 & 92.09 & 84.37 & 95.39 \\ 
            & 15*15 & 76.09 & 70.25 & 22.69 & 95.22 \\ 
    \hline
    Digital & N/A & 93.31 & 89.11 & 78.47 & 94.27 \\ 
    \hline
 \multirow{4}{*}{Ours} & 1*1 & 94.13 & 90.94 & 82.68 & 93.15 \\ 
             & 3*3 & 94.97 & 92.01 & 85.05 & 94.77 \\ 
            & 11*11 & 95.79 & 92.91 & 85.33 & 95.97 \\ 
             & 15*15 & 96.05 & 93.17 & 87.41 & 95.19 \\ 
\hline 
\end{tabular}
\text{Model FLOPs and digital FLOPs unit is MMacs.}
\end{table}

\begin{table}
\centering
% \begin{center}  
\caption{Segmentation performance on car dataset after model compression}
% \end{center}
% \centering
\label{tab:car_after}
\begin{tabular}{lcccccc} 
\hline Model & Kernel size & 1st conv FLOPs (\%) & Model FLOPs & Digital FLOPs & Test (mIoU) \\ \hline 
\multirow{3}{*}{ExtremeC3} & 3*3 & 11.33 & 191.32 & 191.32   & 91.36 \\ 
   & 11*11 & 63.21 & 461.07 & 461.07   & 92.45 \\ 
   & 15*15 & 76.16 & 711.55 & 711.55  & 70.01 \\ 
\hline
 Digital & N/A & N/A & 166.03 & 166.03  & 88.97 \\ 
\hline 
\multirow{3}{*}{Ours} & 1*1 & 3.17 & 174.25 & 166.03   & 90.94 \\ 
 & 3*3 & 11.33 & 191.32 & 166.03   & 94.25 \\ 
 & 11*11 & 60.81 & 423.74 & 166.03  & 95.32 \\ 
 & 15*15 & 64.45 & 467.09 & 166.03  & 93.05 \\ 
\hline 
\end{tabular}
\text{Model FLOPs and digital FLOPs unit is MMacs.}
\end{table}

\subsection{Ablation studies}

\begin{figure}[h]
\begin{center}
\includegraphics[width=0.95 \linewidth]{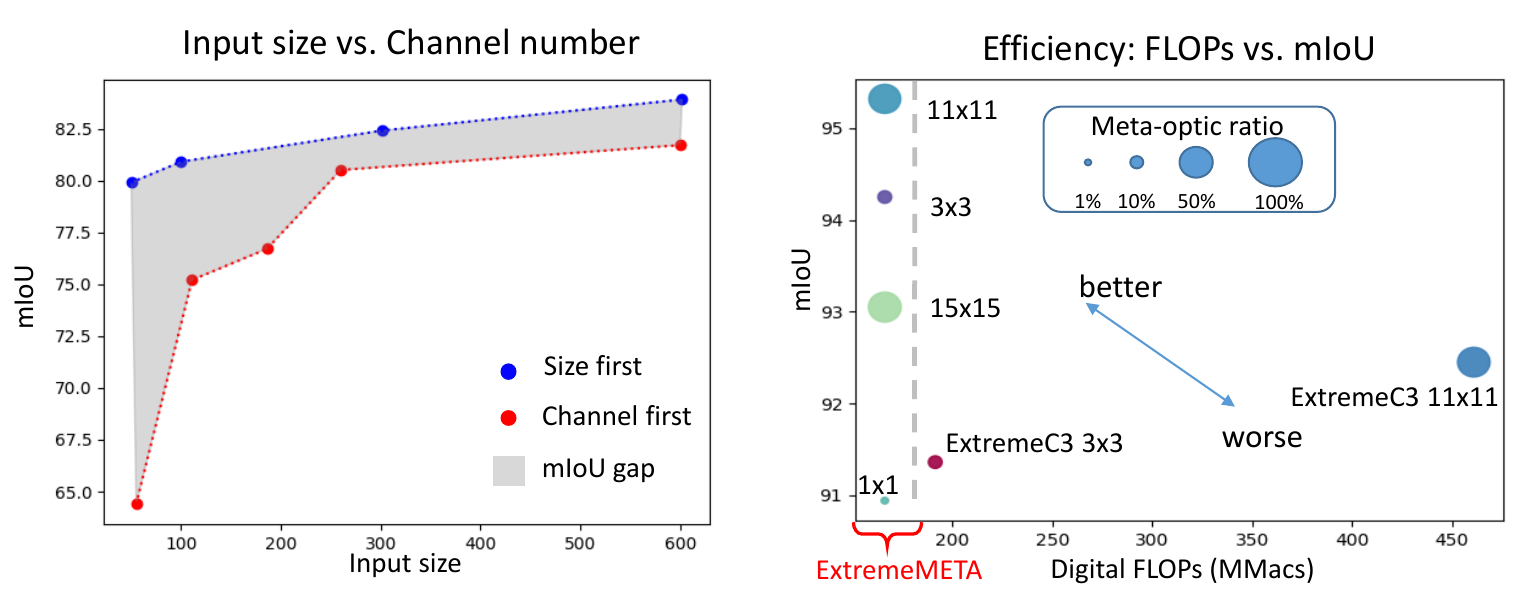}
\end{center}
   \caption{Model ablation study. Left panel: trade-off between input image size and channel number of convolution layer. Right panel: model efficiency visualization comparing model FLOPs and mIoU.}
\label{fig:visual1}
 \end{figure}

Due to the fabrication limitation of the meta-lens array, the priority of channel number and input image size need to be decided. The results of the experiment are shown in Figure ~\ref{fig:visual1}. The left panel illustrates how increasing the input image size enhances performance compared to expanding the number of channels in a convolution layer. The gray area highlights the performance disparity in terms of mean Intersection over Union (mIoU). On the right panel, the effectiveness of utilizing large convolution kernels is assessed. Circles of various colors represent different convolution layer architectures, with the area of each circle indicating the ratio of FLOPs (Floating Point Operations per Second) for the layer when implemented using meta-optic materials. The x-axis represents the model's FLOPs, excluding the layer intended for fabrication.

\subsection{Model compression}
\begin{figure}[h]
\begin{center}
\includegraphics[width=0.8 \linewidth]{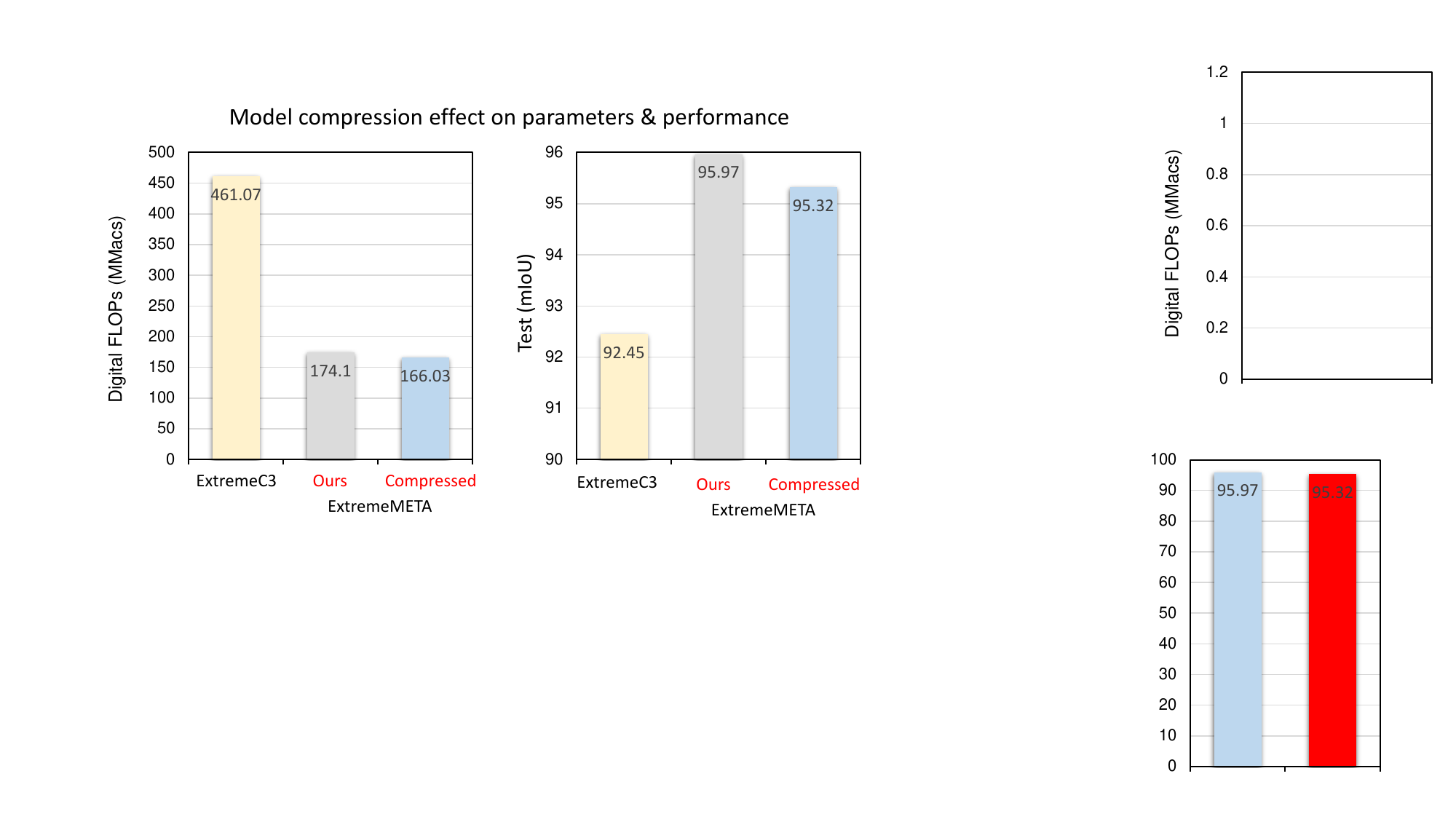}
\end{center}
   \caption{Model compression performance. Left panel: origin model, ExtremeMETA, and compressed model parameters comparison; right panel: model performance after compression.}
\label{fig:visual2}
 \end{figure}

Figure~\ref{fig:visual2} demonstrates that the compressed model achieves a reduction of 8 MMacs in FLOPs, decreasing from 174.10 MMacs to 166.03 MMacs. The right panel indicates that the compressed model maintains equivalent performance to the original model. This consistency in performance illustrates that our ExtremeMETA not only enhances the efficiency of the digital components but also contributes to the overall optimization of the hybrid system.

\section{Discussion}
Given the demonstrated superior performance of large convolution kernels in tasks such as image classification and segmentation, there exists substantial potential for their application in a wider array of complex computer vision tasks. Large convolution kernels have shown remarkable effectiveness in tasks like image classification and segmentation, primarily due to their ability to capture more extensive spatial information and intricate patterns within images. This success suggests that employing large convolution kernels in other computer vision tasks could yield significant improvements.

One such task is object detection, where accurately identifying and localizing objects within images is crucial. By utilizing large convolution kernels, the model can better discern the detailed features of objects, leading to more precise detection results. This can be particularly beneficial in scenarios with small or occluded objects, where finer details are essential for accurate recognition as the results shown in the experiments on the car dataset.

Furthermore, in tasks involving image generation or synthesis, such as style transfer or super-resolution, large convolution kernels can enhance the model's ability to capture intricate textures and details, resulting in more realistic and high-fidelity output images. These kernels can effectively extract and preserve fine-grained features, which are instrumental in faithfully replicating the characteristics of the input images.

The application can be extended to video processing tasks like action recognition or video segmentation, large convolution kernels can enhance the model's capability to analyze temporal and spatial dependencies across frames. By incorporating information from a broader context, these kernels enable a more robust understanding of dynamic scenes, leading to improved performance in tasks requiring temporal coherence and contextual understanding.

The adoption of large convolution kernels holds promise for advancing various complex computer vision tasks beyond traditional image classification and segmentation. Their ability to capture intricate details and spatial relationships makes them a valuable tool for enhancing the performance and capabilities of computer vision models across diverse applications.

\section{Conclusion}
In conclusion, we have introduced a novel large kernel lightweight segmentation model that harnesses the efficiency advantages of optical signal computation while integrating digital processing model compression techniques to further enhance segmentation efficiency. Our model offers larger receptive fields tailored for segmentation tasks on large images, extending its applicability to various vision tasks including image classification, segmentation, and detection. Through extensive evaluations on diverse datasets, including the portrait, Stanford, and KITTI datasets, our proposed approach has demonstrated superior segmentation accuracy compared to state-of-the-art models. Our contributions encompass the introduction of a novel large convolution kernel CNN network for larger reception fields, reduced energy consumption, and lower latency, alongside the introduction of model re-parameterization and sparse convolution kernel compression mechanisms to enhance model performance and efficiency in digital processing. By explicitly addressing task limitations and conducting segmentation tasks on multiple datasets from different categories, our work represents a significant step forward in the development of efficient and effective segmentation models for a wide range of computer vision applications.

\section*{Acknowledgment}

YH and QL acknowledge support from NIH under contract R01DK135597. YH is the corresponding author.
BTS and JGV acknowledge support from DARPA under contract HR001118C0015, NAVAIR under contract N6893622C0030 and ONR under contract N000142112468.
Meta-optic devices were manufactured as part of a user project at the Center for Nanophase Materials Sciences (CNMS), which is a US Department of Energy, Office of Science User Facility, Oak Ridge National Laboratory.

% \ifCLASSOPTIONcaptionsoff
%   \newpage
% \fi
\ifCLASSOPTIONcaptionsoff
  \newpage
\fi



\bibliographystyle{IEEEtran}


% \end{thebibliography}

\bibliography{jist_lmnn}
% \end{thebibliography}

\end{document}